\documentclass[conference]{IEEEtran}
\IEEEoverridecommandlockouts
\usepackage[T1]{fontenc}
\usepackage[utf8]{inputenc}
\usepackage{cite,amsmath,amssymb,graphicx,booktabs,array,tabularx}
\usepackage{algorithm,algpseudocode}
\algrenewcommand{\algorithmicrequire}{\textbf{Input:}}
\algrenewcommand{\algorithmicensure}{\textbf{Output:}}
\algrenewcommand{\algorithmicindent}{1.25em}
\algrenewcommand{\alglinenumber}[1]{\footnotesize #1:}
\algnewcommand{\algorithmicparallelfor}{\textbf{parallel for}}
\algdef{SE}[PAR]{ParallelFor}{EndParallelFor}[1]
  {\algorithmicparallelfor\ #1\ \algorithmicdo}{\algorithmicend\ \algorithmicfor}
\usepackage{tikz,pgfplots}
\usepackage[hyphens]{url}
\usepackage{xurl}
\usepackage[hidelinks]{hyperref}
\usepackage{microtype}
\usepackage{placeins}
\usetikzlibrary{arrows.meta,positioning,calc}
\pgfplotsset{compat=1.17}
\definecolor{inkblue}{RGB}{42,97,130}
\definecolor{mutedgreen}{RGB}{71,123,103}
\definecolor{mutedamber}{RGB}{173,124,52}
\definecolor{mutedred}{RGB}{153,77,82}
\newcolumntype{Y}{>{\raggedright\arraybackslash}X}
\newcommand{\sys}{LumoTree}
\begin{document}
\raggedbottom
\title{LumoTree: Path-Parallel Speculative Verification for Hybrid Language Models}
\author{\IEEEauthorblockN{Zhiyuan Ma}
\IEEEauthorblockA{\texttt{zhiyuanma0520@gmail.com}}
}
\maketitle
\begin{abstract}
Tree speculative decoding for recurrent-hybrid language models requires each accepted path to maintain consistent recurrent state, convolution history, and attention caches. We present LumoTree, a GPU serving design that coordinates verification and commitment through a shared tree descriptor. The verifier processes independent paths in parallel and reuses recurrent state tiles across local updates. Boundary states connect dependent paths. Accepted-path replay publishes the selected continuation into native running state, alongside coordinated convolution gathering and attention-cache remapping. Fused selection, device-resident acceptance, graph replay, and tree-aware attention integrate this organization into the serving cycle. The design separates temporary branch computation from persistent request state while retaining native prefix-cache interfaces. We evaluate numerical behavior, serving performance, and coding-agent outcomes on NVIDIA DGX Spark.
\end{abstract}


\section{Introduction}
Speculative decoding reduces serial target-model invocations by verifying draft tokens in parallel. Its sampling guarantee assumes that verification supplies the intended target probabilities. Tree speculation expands the candidate set, but it does not relax that assumption~\cite{leviathan2023fast,chen2023sampling,miao2023specinfer}. For recurrent-hybrid models, realizing the verifier requires more than an attention mask: a candidate must inherit the state of its ancestors, and the next serving iteration must inherit exactly the path selected for continuation~\cite{yang2024gated,wu2025stree}.

\sys{} executes recurrent paths in parallel, reuses state tiles across updates within each path, and coordinates the accepted continuation across recurrent, convolution, and attention state. An ancestry-aware attention path and a device acceptance walk use the same logical tree as recurrent verification and state publication. A shared descriptor carries ancestry and logical-to-physical mappings through each stage. This system boundary matters because a verifier can compute plausible candidate logits yet publish the wrong state for the next iteration.

The central execution choice is to verify branches in temporary state and replay only the selected path into the native running state. The verifier scans branch paths, while the fixed-shape committer replays accepted operands through native GDN updates in a captured graph. Their finite-precision agreement is a separate requirement, not assumed from the recurrence. This choice exposes concrete optimization questions: how much branch state to cache, when to recompute ancestry, how to organize independent GPU work, and how to keep full-vocabulary probability arithmetic on the device. The attention layout and accepted-state mappings constrain those changes as much as the recurrent equation does.

\begin{figure*}[t]
\centering
\resizebox{\textwidth}{!}{
\begin{tikzpicture}[x=1cm,y=1cm,font=\sffamily\fontsize{8}{10}\selectfont,
  flow/.style={-{Latex[length=1.8mm]},draw=inkblue,line width=0.8pt},
  edge/.style={draw=inkblue!75,line width=0.7pt},
  token/.style={circle,draw=inkblue,fill=white,minimum size=5.5mm,inner sep=0pt},
  chosen/.style={token,draw=mutedgreen,fill=mutedgreen!12,line width=1pt},
  title/.style={anchor=west,font=\sffamily\bfseries\fontsize{9}{11}\selectfont},
  note/.style={font=\sffamily\fontsize{7.5}{9}\selectfont,align=center},
  state/.style={draw=mutedgreen!70,fill=mutedgreen!5,rounded corners=1.5pt,
    minimum height=0.37cm,text width=4.1cm,align=left,inner xsep=5pt}]

\node[title] at (0,4.4) {(a) Propose a candidate tree};
\node[title] at (6.0,4.4) {(b) Verify branch-local computation};
\node[title] at (12.3,4.4) {(c) Select a path and commit state};
\draw[black!18] (0,4.12)--(5.05,4.12);
\draw[black!18] (6.0,4.12)--(11.15,4.12);
\draw[black!18] (12.3,4.12)--(17.5,4.12);

\node[token,fill=black!5,draw=black!45] (root) at (0.55,2.65) {$m$};
\node[token] (a) at (1.8,3.25) {$a$};
\node[token] (b) at (1.8,2.0) {$b$};
\node[token] (c) at (3.1,3.65) {$c$};
\node[token] (d) at (3.1,2.85) {$d$};
\node[token] (e) at (3.1,2.0) {$e$};
\node[token] (f) at (4.35,3.65) {$f$};
\foreach \from/\to in {root/a,root/b,a/c,a/d,b/e,c/f}
  \draw[edge] (\from)--(\to);
\node[note,anchor=north] at (0.55,2.2) {materialized\\prefix};
\node[note,text width=4.7cm] at (2.55,1.05)
  {MTP proposals and suffix extensions\\share one logical ancestry.};
\node[note,text=black!65] at (2.55,0.33)
  {Draft tokens carry proposal probabilities $q$.};
\draw[flow] (4.95,3.0)--(5.8,3.0);

\node[draw=inkblue!60,fill=inkblue!4,rounded corners=2pt,
  minimum width=5.15cm,minimum height=1.97cm] (forward) at (8.575,3.0) {};
\node[font=\sffamily\bfseries\fontsize{8}{10}\selectfont] at (8.575,3.67)
  {Ancestry-aware target forward};
\node[anchor=west] at (6.3,3.23) {Recurrent state};
\node[anchor=east,text=inkblue] at (10.85,3.23) {path scans};
\node[anchor=west] at (6.3,2.79) {Convolution};
\node[anchor=east,text=inkblue] at (10.85,2.79) {path-local filtering};
\node[anchor=west] at (6.3,2.35) {Attention};
\node[anchor=east,text=inkblue] at (10.85,2.35) {tree mask + KV map};
\draw[black!10] (6.3,3.02)--(10.85,3.02);
\draw[black!10] (6.3,2.58)--(10.85,2.58);
\node[draw=mutedamber,fill=mutedamber!6,dashed,rounded corners=2pt,
  text width=4.7cm,align=center,minimum height=0.65cm] (scratch) at (8.575,1.0)
  {Temporary cut states and replay operands};
\draw[flow,draw=mutedamber] (forward.south)--(scratch.north)
  node[midway,right,note,text=black!65] {retain within the step};
\node[note,text=black!65] at (8.575,0.33)
  {Unselected branch state is not published.};
\draw[flow] (11.3,3.0)--(12.1,3.0)
  node[midway,above=2pt,note] {$p$};

\node[chosen,fill=black!5,draw=black!45] (mout) at (12.65,3.05) {$m$};
\node[chosen] (aout) at (13.72,3.05) {$a$};
\node[chosen] (dout) at (14.79,3.05) {$d$};
\node[token,dashed,draw=mutedamber,fill=mutedamber!6] (pending) at (16.55,3.05) {$b^*$};
\draw[flow,draw=mutedgreen] (mout)--(aout);
\draw[flow,draw=mutedgreen] (aout)--(dout);
\draw[flow,dashed,draw=mutedamber] (dout)--(pending);
\node[note] at (13.72,3.67) {Select using $p$ and $q$};
\node[note,text=black!65] at (16.55,3.67) {Pending token};
\node[note,text=black!65] at (16.55,2.51) {not yet in state};
\node[state] (rcommit) at (14.9,1.85) {Recurrent: replay accepted path};
\node[state] (ccommit) at (14.9,1.27) {Convolution: gather history};
\node[state] (kcommit) at (14.9,0.69) {Attention: remap KV entries};
\draw[flow,draw=mutedgreen] (13.72,2.72)--(13.72,2.13);
\draw[-{Latex[length=1.5mm]},dashed,draw=mutedamber,line width=0.7pt]
  (scratch.east)--(11.65,1.0)--(11.65,1.85)--(rcommit.west);
\node[note,text=mutedgreen!80!black,font=\sffamily\bfseries\fontsize{8}{10}\selectfont]
  at (14.9,0.08) {One coherent continuation: $\mathcal{S}(mad)$};

\draw[black!18] (0,-0.35)--(17.5,-0.35);
\node[note,text width=17.2cm] at (8.75,-0.72)
  {\textbf{Shared tree descriptor:} parents, positions, attention visibility, and logical-to-physical slots stay consistent across all three stages.};
\end{tikzpicture}}
\caption{Tree verification and accepted-path commitment. (a) Drafting provides a candidate tree and proposal probabilities $q$. (b) Branch-local computation produces target probabilities $p$ and retains temporary state. (c) Sampling selects a materialized path (here $a,d$); recurrent replay, convolution gathering, and KV remapping publish $\mathcal{S}(mad)$. The newly sampled token $b^*$ remains pending. One descriptor keeps logical ancestry and physical storage aligned.}
\label{fig:pipeline}
\end{figure*}

The paper studies three coupled design questions in recurrent-hybrid tree verification. \textit{First}, how can branch-local computation produce one coherent continuation across recurrent, convolution, and attention state? We define a shared logical tree and coordinate state publication through its logical-to-physical mapping. \textit{Second}, how should verification trade temporary state storage against repeated computation? We use path-parallel scans with shared boundary states and replay the accepted path into persistent state, building on established chain-scheduling and replay mechanisms~\cite{wang2026specla,oda2026marginals}. \textit{Third}, how can this organization reduce serving overhead? We combine fused candidate selection, device-resident acceptance, and grouped split-K attention with a spine-contiguous KV layout. We evaluate component numerical agreement, recorded-request replay against native decoding, and an exploratory coding-agent deployment. Section~\ref{sec:limitations} bounds these results.

The target workload is a long-running coding agent: generation alternates with repository inspection, tool execution, and tests, and a task can require many model requests. Performance therefore depends on useful completed work as well as token production. Our workload evaluation uses explicitly identified Astropy tasks from SWE-bench Verified~\cite{jimenez2024swe}. We report the model, serving route, task cohort, and measurement definition together; a decode-rate change is not a task-completion speedup. Candidate-selection and attention tests complement the agent-workload measurements. Penalty processing builds each verification row's history from its own root-to-node path; a captured-row audit and a Monte-Carlo test of the acceptance walk find no departure from the autoregressive target beyond the top-$k$ tie convention (Section~\ref{sec:sampling-validation}).

Tree verification and compact accepted-state reconstruction have direct prior art. Section~\ref{sec:related} positions our implementation against those mechanisms. Sections~\ref{sec:contract}--\ref{sec:kernel} develop the architecture and kernel design; Section~\ref{sec:optimization} describes the optimizations that connect them to the serving cycle. Appendix~\ref{app:audit} gives configuration and measurement details.

\section{Background and Related Work}
\label{sec:related}
\subsection{Drafting, verification, and serving}
Speculative sampling constructs an exact target sample from proposal and residual distributions, conditional on the probabilities actually supplied~\cite{leviathan2023fast,chen2023sampling}. SpecInfer supplies the tree-verification framing; Sequoia, EAGLE-2, and OPT-Tree investigate tree construction and its performance consequences~\cite{miao2023specinfer,chen2024sequoia,li2024eagle,wang2024opt}. EAGLE-3 and DFlash further change the drafter, through feature-conditioned autoregression and block-parallel diffusion proposals respectively~\cite{li2025eagle3,chen2026dflash}. These drafting mechanisms are distinct from the recurrent verifier and state-publication choices studied here.

PagedAttention and SGLang address the scheduling, memory, and execution substrate~\cite{kwon2023efficient,zheng2024sglang}. Marconi addresses prefix-cache admission and eviction for hybrid models~\cite{pan2024marconi}. Hybrid serving additionally manages recurrent-state pools alongside KV storage, as documented by SGLang's hybrid-model integration~\cite{pytorch2025sglang,sglang2026qwen}. Thus neither a fast isolated recurrent kernel nor low transient state memory alone determines user-visible latency. Flash-Decoding splits the KV sequence and combines partial attention results; FlashInfer develops customizable attention with head/query fusion and split-KV execution; DeFT specializes KV grouping and partitioning to trees~\cite{dao2023flashdecoding,ye2025flashinfer,yao2024deft}. FastTree groups queries with shared contexts to reuse KV tiles and partitions tree work to control execution cost~\cite{pan2025fasttree}. These attention methods establish precedents for on-chip reuse and tree-aware partitioning. Our modified FlashAttention-2 kernel combines ancestry-aware decode, grouped split-K execution, and a physical layout coordinated with accepted-path publication~\cite{dao2023flashattention}.

\subsection{Recurrent and hybrid speculation}
Gated DeltaNet augments a decayed recurrent state with a data-dependent delta update~\cite{yang2024gated}. Parallel delta-rule formulations provide relevant algebraic background~\cite{yang2024delta}. Snakes and Ladders introduces activation replay for SSM speculative decoding, and STree also uses replay to recover continuation state~\cite{wu2024snakes,wu2025stree}. Stateful speculation is established territory: STree studies speculative trees for hybrid state-space models, SpecMamba targets an FPGA implementation, and component-aware self-speculation studies how hybrid components affect the design~\cite{wu2025stree,zhong2025specmamba,borobia2026component}. These works motivate an explicit state lineage; they do not justify treating every recurrent update or numerical implementation as interchangeable.

\subsection{Direct GDN comparisons}
\textbf{Trees from Marginals (Weaver)} combines a factorized drafter with an autoregressive adapter. Its GDN verifier uses an ancestor-masked triangular solve and replays the selected path through a short recurrence at commit time; padded verification shapes support CUDA graphs~\cite{oda2026marginals}. Thus read-only speculative verification followed by accepted-path replay is direct prior art, even when the verification arithmetic differs from ours.

\textbf{SpecLA} keeps value tiles of recurrent state on chip across updates within a chain and schedules independent ready chains in parallel. Boundary states connect dependent chains. This directly overlaps both the state reuse and path scheduling in \sys{}. SpecLA buffers accepted factors for a later fused update, whereas our publication stage replays recorded operands through native GDN updates into running rows. Its evaluation uses a pure GDN model; our serving boundary additionally coordinates convolution and softmax attention~\cite{wang2026specla}. SpecLA's token-replay baseline regenerates factors, so its cost cannot be assigned to our replay of cached operands.

\textbf{Bole} derives a tree-structured closed form and evaluates it with a value-tiled finite-polynomial solver. It reuses intermediate tiles on chip, stores compact update factors, and reconstructs the accepted state. Its SGLang integration combines batch-wide verification budgeting with captured GPU execution. Evaluation includes unquantized Qwen3.5-27B on GB10 and replay of recorded OpenHands agent sessions~\cite{wang2026bole}. This is a direct hybrid-serving comparison. Our evaluation reports recorded-request replay separately from running the coding agent on tasks.

Concretely, Bole's correction system is $(I+G)U=R$, where $G$ contains strict-ancestor interactions and $R$ is the source correction. Tree depth $d$ gives $G^{d+1}=0$, so $U=\sum_{m=0}^{d}(-G)^mR$. This finite identity changes the arithmetic schedule from sequential updates to matrix propagation~\cite{wang2026bole}. Our path preserves sequential updates and replays accepted operands; Section~\ref{sec:reconstruction-numerics} explains why the numerical comparison remains an empirical question.

\textbf{TreeWY} expresses GDN tree verification as a triangular system and reconstructs the accepted state from pseudo-values. The author implementation fuses reconstruction of the previous accepted state with the next verification, reusing a resident state tile. Its GDN tree kernel supports CUDA graph capture; the paper's wider-tree fallback to piecewise execution concerns ancestry masks in the softmax-attention integration~\cite{ghantasala2026treewy,ghantasala2026treewycode}. Its B200 evaluation uses greedy verification and disabled prefix caching. A comparison must therefore include fused commit work and distinguish kernel support from whole-model execution policy.

\textbf{ReplaySSM} caches recurrent inputs and delays state materialization. Its GDN path uses corrected inputs from a triangular solve, with commitment represented through buffer progress and later state updates~\cite{replayssm2026}. \textbf{OneLA} addresses a related state-sharing problem in recommendation beam search: compact transition records and ancestry indices allow projection computation without reconstructing a full state for each beam, and a fused kernel shares the prompt-derived state across beams~\cite{yang2026onela}. Its beam-search output and commitment policy differ from speculative acceptance. Table~\ref{tab:related} compares state lifetimes; published speed ratios from different models and workloads are not combined into a ranking.

\begin{table}[t]
\caption{Recurrent verification and continuation policies. These mechanism comparisons do not rank the systems by performance.}
\label{tab:related}
\centering\footnotesize
\begin{tabularx}{\columnwidth}{@{}>{\raggedright\arraybackslash}p{0.22\columnwidth}YY@{}}
\toprule
Method & Verification work & Continuation state \\
\midrule
Weaver~\cite{oda2026marginals} & Ancestor-masked triangular solve & Short selected-path recurrence \\
SpecLA~\cite{wang2026specla} & Resident state tiles; parallel chains & Buffered factors; later fused update \\
Bole~\cite{wang2026bole} & Finite-Neumann correction solve & Compact factors; state reconstruction \\
TreeWY~\cite{ghantasala2026treewycode} & Triangular solve in fused kernel & Prior-path reconstruction with next verify \\
ReplaySSM~\cite{replayssm2026} & Corrected inputs; cached transitions & Deferred state update \\
OneLA~\cite{yang2026onela} & Shared-base projections across beams & Compact per-beam transition records \\
\sys{} & Resident state tiles; parallel paths & Native replay; coordinated convolution and KV publication \\
\bottomrule
\end{tabularx}
\end{table}

\subsection{Design focus}
\sys{} connects path-parallel recurrent computation to one accepted continuation across the hybrid model. The shared descriptor links path boundaries, convolution ancestry, attention visibility, physical KV slots, and the next drafter input. Native replay preserves the serving system's recurrent-state ownership while selected-path gathering and remapping publish the other state components. Spine-first KV placement controls the physical attention reduction layout within this design, building on established floating-point reduction and batch-invariance concerns~\cite{he2025determinism}. Evaluation separates component behavior from complete agent-workload performance.

\section{Tree-Verifier Architecture}
\label{sec:contract}
\subsection{State and the execution boundary}
Let $m$ be a materialized token prefix: the tokens already consumed by the target's forward computation. Write its persistent state as
\begin{equation}
\mathcal{S}(m)=\bigl(R(m),C(m),K(m)\bigr),
\end{equation}
where $R$ collects recurrent matrices, $C$ collects convolution histories, and $K$ collects attention KV entries. The logical emitted prefix can also include a newly sampled token $b$ that has not yet been consumed. The continuation boundary is then $(\mathcal{S}(m),b)$, not a claim that $\mathcal{S}(mb)$ has already been computed. This distinction is essential when counting the correction or bonus token while checking cache state.

A tree descriptor contains candidate token IDs, parent indices, depth/position information, attention visibility, and maps between logical nodes and physical slots. For a candidate node $i$, each target layer must consume the state obtained from the candidate's root-to-parent path. At publication, the selected materialized path, pending-token convention, and next-drafter input row must agree. Figure~\ref{fig:contract} illustrates the three state surfaces coordinated by the verifier.

\subsection{The recurrence and path locality}
For one GDN head, let $R\in\mathbb{R}^{d_v\times d_k}$. Queries and keys are in $\mathbb{R}^{d_k}$ and values in $\mathbb{R}^{d_v}$. With decay $\alpha_i$ and write strength $\beta_i$, a node applies the standard update~\cite{yang2024gated}
\begin{align}
R'_i &= \alpha_i R_{\pi(i)}, &
\delta_i &= \beta_i\bigl(v_i-R'_i k_i\bigr),\\
R_i &= R'_i+\delta_i k_i^\top, &
o_i &= R_iq_i,
\end{align}
where $\pi(i)$ is its parent. A previous node in the packed buffer need not be $\pi(i)$. The causal convolution likewise requires path predecessors rather than adjacent packed rows. The attention visibility relation must select the same logical prefix and ancestry. These requirements follow from the model's computation; a correct attention mask cannot repair a wrong recurrent parent.

The implementation verifies nodes with branch-local scratch and makes the selected continuation durable through replay and remapping. This is an implementation route, not the only way to satisfy the contract. In particular, rejecting a branch requires that its contents cannot influence future computation; it does not require zeroing every scratch byte if future reads are correctly restricted.

\subsection{A descriptor shared by verification and commitment}
Figure~\ref{fig:pipeline} shows how a shared descriptor connects drafting to continuation. Native multi-token prediction (MTP) supplies a principal chain and runner-up candidates; suffix prediction extends selected paths within a bounded candidate capacity~\cite{oliaro2025suffix,arctic2026suffix}. Combining retrieval and neural drafting has prior art, including SAM Decoding~\cite{hu2024sam}. The verifier derives attention visibility, physical packing, path scheduling, and the acceptance walk from the same logical topology. Padding permits reusable execution shapes while remaining invisible to attention and selection. The particular tree geometry used in our experiments is specified in Section~\ref{sec:protocol}.

\begin{figure}[t]
\centering
\begin{tikzpicture}[font=\scriptsize,
 box/.style={draw=inkblue,fill=inkblue!7,rounded corners,align=center,minimum height=0.7cm},
 arr/.style={-{Latex[length=1.7mm]},thick,draw=inkblue}]
\node[box,minimum width=2.0cm] (prefix) at (0,0) {Materialized\\prefix $m$};
\node[box,minimum width=1.8cm] (tree) at (2.7,0) {Verify\\candidate tree};
\node[box,draw=mutedgreen,fill=mutedgreen!8,minimum width=2.0cm] (path) at (5.4,0) {Select accepted\\path $a$};
\draw[arr] (prefix)--(tree);\draw[arr] (tree)--(path);
\node[box,minimum width=1.8cm] (r) at (0,-1.5) {Recurrent\\$R(ma)$};
\node[box,minimum width=1.8cm] (c) at (2.7,-1.5) {Convolution\\$C(ma)$};
\node[box,minimum width=1.8cm] (k) at (5.4,-1.5) {Attention KV\\$K(ma)$};
\draw[arr] (path.south) -- ++(0,-0.35) -| (r.north);
\draw[arr] (path.south) -- ++(0,-0.35) -| (c.north);
\draw[arr] (path.south)--(k.north);
\node[align=center,text width=7cm] at (2.7,-2.45) {Same selected materialized prefix on all three surfaces;\\correction token $b$ may remain pending for the next forward.};
\end{tikzpicture}
\caption{The accepted-prefix interface. Recurrent, convolution, and attention-KV state must all describe the same materialized prefix. A newly sampled correction or bonus token can remain pending until the next forward.}
\label{fig:contract}
\end{figure}

The attention layout places the principal spine first in the physical suffix and permutes KV write locations and ancestry-mask key columns while retaining logical query-row and position order. Selection remains in logical-node space; accepted-path publication maps those nodes back to their physical slots. This distinction prevents a layout optimization from silently changing which key/value entries a branch reads. Recurrent paths begin at the native pre-tree state or a transient cut state; the device committer publishes the selected recurrent and convolution state into native running rows and re-linearizes its attention KV entries.

\begin{algorithm*}[t]
\caption{Tree verification and native continuation}
\label{alg:verify}
\normalsize\linespread{1.08}\selectfont\raggedright
\begin{algorithmic}[1]
\Require Native continuation state; candidate tree $T$; proposal probabilities $q$
\Ensure Emitted tokens; updated native state; next drafter input; any pending token
\Statex \textbf{Verify candidates}
\State Map $T$ to physical slots, positions, and ancestor masks.
\State $p\gets$ target probabilities from path-local GDN, convolution, and attention computation.
\State Retain candidate hidden states, rounded replay operands, convolution histories, and temporary KV entries.
\Statex \textbf{Select one continuation}
\State $(A,b^*)\gets$ accepted materialized path and pending token from sampling with $p$ and $q$.
\Statex \textbf{Publish the same path across all state components}
\For{each GDN layer}
  \State Replay recorded operands along $A$ into the native recurrent state.
  \State Gather the convolution history for $A$ into the native running row.
\EndFor
\State Remap the accepted KV entries using the same logical-to-physical slot map.
\State Select the matching hidden-state row for the next drafter input.
\State Advance positions along $A$; keep $b^*$ pending until its target forward.
\State \Return Emitted tokens and the updated continuation boundary.
\end{algorithmic}
\end{algorithm*}

Algorithm~\ref{alg:verify} states the intended probability and continuation contract and separates candidate verification, path selection, and state publication. History-dependent logits processing builds each verification row's penalty history from its own root-to-node path; Section~\ref{sec:sampling-validation} audits the deployed rows. Every publication operation uses the same materialized path $A$. A newly sampled token $b^*$ can be emitted while remaining pending: its state is created by the next target forward.

\section{GPU Scan and Accepted-Path Replay}
\label{sec:kernel}
\subsection{Path-parallel verification and state lifetime}
The GDN verifier decomposes the tree into dependent path groups, a scheduling family also studied by SpecLA~\cite{wang2026specla}. A first path starts from the request's pre-tree state and exports the cut states needed by downstream paths. The second level evaluates independent branch paths from those cut states. Each GPU program covers one path, head, and value tile, retaining the full key dimension in its working state tile. Sequential updates reuse this tile through the path loop rather than loading a full recurrent state for every node. Candidate outputs continue through the model; rounded operands are recorded for later accepted-path replay. The source expresses this local reuse; register allocation and spills remain compiler- and geometry-dependent.

Cut states and operand rings are temporary within a verification step. The durable continuation resides in the request-owned native recurrent and convolution running rows. Replay starts from the pre-tree running state and publishes the accepted frontier back to those rows; prefix-cache snapshots and restores use that native layout. Static work buffers can be reused by CUDA graphs, but they do not replace the request's persistent state with a branch-leaf pointer.

The two levels execute in separate GPU launches and exchange only the boundary states required by downstream paths. This schedule exposes parallel branch work while retaining the state-transfer cost at the decomposition boundary.

\begin{algorithm*}[t]
\caption{Path-parallel scan and accepted-path replay}
\label{alg:scan-replay}
\normalsize\linespread{1.08}\selectfont\raggedright
\begin{algorithmic}[1]
\Require Candidate tree $T$ and the pre-tree recurrent state for each GDN layer
\Ensure Recurrent outputs for all candidates; native recurrent state for the accepted path
\Statex \textbf{Verify paths; keep branch state temporary}
\For{each GDN layer reached during the target forward}
  \State Compute and record this layer's rounded node operands.
  \For{each path group, in dependency order}
    \ParallelFor{each ready path, head, and value tile}
      \State Load the working state tile from the pre-tree state or a saved parent cut state.
      \For{each node on this path, in ancestry order}
        \State Update the working tile with this node's operands; write its recurrent output.
        \State Save the tile if a downstream path needs this node's state.
      \EndFor
    \EndParallelFor
  \EndFor
\EndFor
\Statex \textbf{Finish verification and select a path}
\State Complete the target forward and sampling to obtain the accepted materialized path $A$.
\Statex \textbf{Replay only the accepted path into native recurrent state}
\For{each GDN layer}
  \State Start from this layer's pre-tree recurrent state.
  \For{each node on $A$, in path order}
    \State Apply the native GDN update using this layer's recorded operands.
  \EndFor
  \State Publish the resulting recurrent state to the request's native running row.
\EndFor
\end{algorithmic}
\end{algorithm*}

\subsection{Sequential arithmetic and accepted-path replay}
\label{sec:reconstruction-numerics}
Verification uses sequential path scans with fp32 carried state, raw gate inputs, and in-kernel query/key normalization. Accepted-path replay uses native fused GDN updates with recorded accepted operands and in-place publication to the fp32 state bank. Its graph contains one native update call per GDN layer. Query/key normalization, beta conversion, gate evaluation, and output casts can differ between these implementations; agreement is not guaranteed by their common real-arithmetic recurrence.

The sampling stage produces accepted node indices and path lengths on the GPU. Recurrent replay, convolution publication, and attention-KV publication consume this same path. Graph capture reduces host launch overhead for the sequence of per-layer native updates; each layer retains its own update kernel.

\subsection{Temporary state versus recomputation}
Path decomposition caches the states needed where downstream paths branch, then reuses them across independent work. If a program exports $N_c$ cut states with value-tile width $B_v$ and key dimension $d_k$, their fp32 payload is
\begin{equation}
 M_{\mathrm{cut}}=4N_cB_vd_k\ \text{bytes}.
\end{equation}
This counts one head and value tile's logical export payload. The current reusable scratch allocation reserves a full matrix slot for each actual node even though only required cut states are written; allocated bytes therefore differ from state traffic. Recomputing ancestry could remove some exports at the cost of repeating updates. Exporting every candidate state would instead trade more writes for less recomputation. The two-level path schedule occupies a particular point in this design space.

Accepted-path replay carries one state tile through a bounded path loop. Its carried-state footprint is $O(B_vd_k)$ in tree size, while the recorded operands and candidate outputs scale with the verification shape. Static padding makes graph buffers and kernel geometry reusable, but padded rows must remain invisible to attention and selection. Neither a shorter logical path nor fewer state writes alone establishes a faster complete request.

\begin{figure}[t]
\centering
\begin{tikzpicture}[font=\scriptsize,
 box/.style={draw=inkblue,fill=inkblue!7,rounded corners,align=center,text width=3.05cm,minimum height=0.7cm},
 arr/.style={-{Latex[length=1.6mm]},thick,draw=inkblue}]
\node[box] (prior) at (0,0) {Pre-tree state $R_0$};
\node[box] (scan) at (0,-1.2) {Scan dependent path groups\\temporary cut-state exports};
\node[box] (out) at (0,-2.4) {Candidate outputs\\complete forward and acceptance};
\node[box,draw=mutedamber,fill=mutedamber!8] (ring) at (3.55,0) {Recorded node operands};
\node[box,draw=mutedgreen,fill=mutedgreen!8] (replay) at (3.55,-1.2) {Replay selected path from $R_0$\\one carried state tile};
\node[box,draw=mutedgreen,fill=mutedgreen!8] (publish) at (3.55,-2.4) {Publish accepted state\\to native running rows};
\draw[arr] (prior)--(scan);
\draw[arr] (scan)--(out);
\draw[arr] (ring)--(replay);
\draw[arr] (prior.east)--(replay.north west);
\draw[arr] (out.east)--node[above,sloped,font=\tiny]{selected path}(replay.south west);
\draw[arr] (replay)--(publish);
\end{tikzpicture}
\caption{State lifetime in the path-scan verifier. Temporary cut states connect path groups; recorded operands support selected-path replay. Only the accepted frontier becomes native persistent state.}
\label{fig:scan-replay}
\end{figure}

The closest scheduling comparison is SpecLA: both systems carry state tiles through sequential paths and parallelize ready paths. Their continuation policies differ. SpecLA buffers accepted factors for application in a later fused update~\cite{wang2026specla}; \sys{} immediately replays recorded operands through native per-layer updates and publishes the matching convolution and attention state. Weaver also uses selected-path replay, but verifies candidates with a triangular solve~\cite{oda2026marginals}. Bole reconstructs accepted states from compact factors in a batched commit~\cite{wang2026bole}. TreeWY can fuse prior-frontier reconstruction with the next verification while retaining its working tile~\cite{ghantasala2026treewycode}. Thus, \sys{} pairs path recurrence with immediate native replay and coordinated hybrid-state publication. This pays explicit accepted-path update work in exchange for returning each iteration to the native state representation. Complete-cycle measurements must include that tradeoff.

\section{Optimizing the LumoTree Verification Cycle}
\label{sec:optimization}
The architectural choice in Section~\ref{sec:kernel} is realized through the implementation changes below. Three representations must agree: logical tree nodes, temporary verification storage, and the native state used by the next forward. \sys{} carries one accepted path across these representations and reorganizes preparation and selection around it, using established techniques such as fusion and graph replay. The nontrivial constraints arise at their boundaries: changing a slot order, a rounding step, or a tied candidate can change which continuation the next iteration consumes. Table~\ref{tab:optimization-map} summarizes the changes and these constraints.

\begin{table*}[t]
\caption{Concrete implementation changes around the path verifier. The third column identifies the coordination or numerical constraint that makes each change nontrivial. Workload rates measure their combined execution.}
\label{tab:optimization-map}
\centering\footnotesize
\renewcommand{\arraystretch}{1.12}
\begin{tabularx}{\textwidth}{@{}>{\raggedright\arraybackslash}p{0.23\textwidth}YY@{}}
\toprule
Starting work or constraint & \sys{} change & What must remain consistent \\
\midrule
Host path-list construction between acceptance and publication & Device-selected path feeds captured native replay, convolution, KV and drafter selection & All consumers select the same materialized prefix; pending and padded tokens do not advance state \\
Logical ancestry does not determine physical KV placement & Spine-contiguous slots with coordinated writes, mask columns and accepted-path remap & Logical query order, physical source slots and native destination positions agree \\
Per-node convolution histories and repeated layer preparation & Static ancestry gathers, fused ordered taps and per-layer operand buffers & Tap order, cast boundaries, history padding and layer identity \\
Separate score preparation and argmax/top-k operations & One shared logits tensor and fused selection into graph buffers & Both reference tie conventions, candidate order and duplicate behavior \\
Repeated KV staging and limited decode parallelism & Grouped query heads and split-K in the patched tree-attention kernel & Tree visibility and slot mapping across partitions; characterized reduction error \\
\bottomrule
\end{tabularx}
\end{table*}

\subsection{One device-selected path drives native continuation}
Selecting valid output tokens does not by itself update a hybrid model's state. The recurrent matrix must contain the accepted updates, convolution must retain the accepted history, attention must reference the accepted KV entries, and the drafter must receive the corresponding hidden row. Reconstructing these choices separately on the host introduces path-list preparation and synchronization between acceptance and publication.

\sys{} replaces that handoff with fixed-shape device products: accepted node indices, path lengths, emitted tokens, and the selected final row. These products feed native recurrent replay, convolution gathering, KV remapping, and drafter-row selection. Recorded operands are staged for the native per-layer updates in one captured committer graph; convolution and KV publication use the same selected path. Graph capture amortizes launch orchestration, while the shared path prevents each state component from independently interpreting acceptance.

The distinction between emitted and materialized tokens must survive this handoff. A correction or bonus token may be emitted while remaining pending for the next target forward; padded path entries must not advance state. After publication, the implementation stores persistent state in native request rows and branch cut states in temporary scratch. The native representation supports the intended prefix-cache and serving-graph interface. The implementation captures the serving, drafter, and committer stages separately.

\subsection{One slot map governs attention and publication}
An ancestry mask specifies visible keys, but does not specify their physical placement or the reduction groups that process them. Interleaving masked branch columns with spine keys can change the online-softmax reduction layout even when logical visibility is unchanged. Physical ordering therefore matters both for addressing and for finite-precision execution~\cite{he2025determinism}.

\sys{} places the principal spine contiguously in the candidate suffix while retaining logical query-row and position order. It applies the same permutation to KV write slots and ancestry-mask key columns. Acceptance remains in logical-node space; publication translates the selected nodes through that slot map into native continuation positions. Reordering only writes, only masks, or only the final copy would leave another consumer reading a different branch. The implementation change is this coordinated mapping across verification and publication.

The patched FlashAttention-2 decode kernel combines this layout with grouped query heads that share staged KV tiles and split-K context partitions. KV reuse and partitioned attention are established techniques~\cite{yao2024deft,pan2025fasttree}; here they are applied under the tree's visibility and slot mapping. They target repeated KV staging and insufficient parallel work at batch one. Prefill retains a native-compatible attention path. The current component tests measure repeatability and numerical error of the deployed attention kernel; the isolated effect of spine placement remains an ablation question.

\subsection{Precompute branch histories while preserving arithmetic}
The starting tree-convolution implementation constructs candidate histories with per-node operations and repeats path-row preparation across layers. \sys{} derives static gather indices from the topology, gathers windows from the prior convolution history and candidate inputs, and fuses the tap computation. Stacked operand buffers retain each GDN layer's inputs for subsequent accepted-path replay.

This transformation must preserve more than ancestry. Convolution tap order and cast boundaries affect the values recorded for recurrent updates, and unused history positions must retain the intended padding. A generic reduction can compute the same real-arithmetic sum in a different floating-point order. The fused implementation therefore retains the ordered tap operations and conversion points while replacing repeated history construction with prepared data movement. Each layer still records and replays its own operands.

\subsection{Fuse draft selection without changing proposals}
At each MTP head depth, \sys{} computes one full-vocabulary logits tensor and reuses it for the principal token and sibling candidates. A fused CUDA kernel writes the spine token and three ordered candidate indices directly into reusable graph buffers, replacing separate argmax/top-k reductions and their output copies. The principal chain advances through captured MTP iterations; sibling selection reuses its scores, and suffix prediction extends selected paths within the fixed verification capacity.

The subtlety is that argmax and top-k do not have interchangeable tie behavior. The reference argmax chooses the lowest maximal token index, while its top-k operation orders selected ties differently. Simply taking the first top-k result as the spine token can change the proposal tree. Our fused selection preserves both reference outputs, including duplicate proposals where produced by the reference. Exact-output and captured-replay tests check this component. The optimization removes redundant score preparation and selection work while retaining the vocabulary projection.

\section{Evaluation Setup}
\label{sec:protocol}
We evaluate Qwen3.8-27B NVFP4 on a single NVIDIA DGX Spark (GB10) using a coding-agent study, recorded-request replay, full-model continuation diagnostics, and isolated recurrent measurements. Qwen Code performs repository edits and tool calls on ten SWE-bench Verified Astropy tasks~\cite{jimenez2024swe}, once per task under each of three serving arms: \sys{}, native five-token MTP, and SGLang EAGLE. The benchmark evaluator tests nonempty patches; empty patches count as unresolved attempts. Separate replays compare serving configurations on two disjoint corpora of 43 requests each. Decode rates pool token counts and decode durations over the same completed requests, excluding time to first output. Agent duration additionally includes tools and repository operations; server startup and evaluation are separate.

All three arms load the same ModelOpt mixed-precision checkpoint. Target MLP weights and the vocabulary projection use NVFP4, attention projections use FP8, and activations and KV storage use BF16; recurrent state uses FP32. The checkpoint's one-layer MTP module remains BF16. Native MTP and \sys{} use this module, and SGLang invokes it through its EAGLE implementation without a separately trained draft checkpoint. Native MTP and \sys{} share the vLLM build; SGLang uses a separate serving stack.

The evaluated tree has 27 draft candidates and one root, padded to 32 verification rows. A five-depth MTP head supplies 15 candidates; suffix prediction adds a six-token principal tail and rescue branches of four and two tokens. Maximum draft depth is eleven. Five MTP passes construct the head, including four forwards after its initial prediction; the deeper suffix extensions do not require additional MTP passes. Attention groups query heads in pairs and divides the context into four partitions. The model has 48 GDN layers and sixteen full-attention layers. All main \sys{} workload runs use this geometry and the patched FlashAttention-2 path. Appendix~\ref{app:settings} gives the serving configuration; the accompanying artifact identifies each source and kernel version.

The first replay corpus contains 43 requests recorded from the Astropy tasks \texttt{12907} and \texttt{13033}. Each configuration receives the same saved messages and tool definitions, sequentially with at most one active request. Reported vLLM prompt lengths range from 23,024 to 68,557 tokens. Sampling uses temperature 0.6, top-$p$ 0.95, top-$k$ 20, and presence penalty 1.0, with a 1,024-token response cap and no per-request seed. This protocol measures fixed-request serving behavior; generated tool calls are not executed. Penalties use per-path histories (Section~\ref{sec:sampling-validation}).

Native autoregressive decoding, chain MTP, and \sys{} use the same checkpoint, vLLM image, tokenizer, chat template, and common serving settings, including prefix caching, chunked prefill, a 131,072-token context limit, and one active sequence. The native arms use their native attention invocation; \sys{} uses its specialized tree-attention path. For the cross-stack comparison, we capture the complete input token sequence from vLLM's tokenizer endpoint and pass those IDs directly to SGLang's generation endpoint. All 43 sequences match the vLLM prompt lengths and the SGLang run records. SGLang EAGLE uses the same checkpoint, FlashInfer attention, and its own prefill and streaming implementation.

The disjoint corpus selects 43 of 61 captured requests from native-MTP agent executions on tasks \texttt{14096}, \texttt{14309}, \texttt{14508}, and \texttt{14995}, across five conversations. Selection allocates requests proportionally across conversations and spaces them through capture order. Its manifest is frozen before any replay; neither message bodies nor initial user messages overlap the first corpus. LumoTree and native MTP-5 replay these saved messages in three runs each; SGLang has one run. SGLang uses its chat endpoint for this corpus, so matching messages does not establish matching serialized input IDs. Input collection is distinct from the 30-attempt task-outcome study.

For vLLM and the disjoint corpus's SGLang chat endpoint, the replay client measures time from the first nonempty streamed content, reasoning, or tool-call delta to stream completion. For the first corpus's SGLang token-ID endpoint, timing begins at the first streamed increase in the completion-token count. We apply Eq.~\ref{eq:pooled-decode} to these client durations and server-reported completion counts, pooling matched token and time totals across requests and repetitions. The estimate excludes first-output latency and server startup. Subsequent stream delivery and parser behavior remain in the measured duration; the endpoint observables are not identical.

\section{Results}
\label{sec:results}
\subsection{SWE-bench Verified coding-agent workload}
\label{sec:agent-workload}
The coding-agent study completes 30 attempts: ten selected Astropy tasks, once each under \sys{}, native MTP-5, and SGLang EAGLE. All arms use Qwen Code 0.19.4, the same checkpoint, a 24,000-token response cap, and a 9,000-second task budget. The agent inspects repositories, edits files, and executes tools. Table~\ref{tab:agent-study} summarizes all attempts, including six empty patches; Appendix~\ref{app:workload} gives task-level results. All arms use presence penalty 1.0; the \sys{} arm applies per-path penalty histories (Section~\ref{sec:sampling-validation}).

For completed request $r$, let $n_r$ be the output-token count, $\ell_r$ the server-recorded end-to-end latency, and $f_r$ the time to first output. We report the pooled decode rate
\begin{equation}
 D_{\mathrm{pool}} =
 \frac{\sum_{r=1}^{R}(n_r-1)}{\sum_{r=1}^{R}(\ell_r-f_r)}.
 \label{eq:pooled-decode}
\end{equation}
Counts and durations cover the same completed requests, including requests that summarize an overlong agent context. The numerator counts output intervals; the denominator excludes first-output latency and between-request agent/tool work. Thus $D_{\mathrm{pool}}$ measures token production over accumulated request durations, rather than aggregate wall-clock throughput.

\begin{table}[t]
\caption{Ten-task coding-agent study, one attempt per task and arm. Minutes sum agent durations, including tools but excluding server startup and evaluation. Decode rates use server-counter differences in Eq.~\ref{eq:pooled-decode}.}
\label{tab:agent-study}
\centering\small
\setlength{\tabcolsep}{3pt}
\begin{tabular}{@{}lrrrr@{}}
\toprule
Arm & Resolved & Requests & Tokens/s & Minutes \\
\midrule
\sys{} & 5/10 & 248 & 26.12 & 190.8 \\
Native MTP-5 & 6/10 & 224 & 27.17 & 99.5 \\
SGLang EAGLE & 5/10 & 210 & 29.89 & 103.0 \\
\bottomrule
\end{tabular}
\end{table}

\sys{} produces eight nonempty patches, three of which fail tests, and two empty patches. Native MTP-5 produces six resolving patches, two failing-test patches, and two empty patches; SGLang produces five, three, and two respectively. Empty patches count as failures without invoking the test evaluator. These are descriptive outcomes on the selected tasks, not benchmark-wide quality estimates.

The live agent trajectories differ: \sys{} emits 258,889 output tokens, compared with 134,643 for MTP-5 and 158,827 for SGLang. Their request counts, prompt lengths, and tool calls therefore differ as well. The lower \sys{} decode rate and longer summed agent duration in this study contrast with the fixed-request replay results below. Sharing tasks and agent settings does not isolate a verifier's effect on task completion.

\subsection{Recorded-request replay}
\label{sec:request-replay}
Table~\ref{tab:request-replay} reports complete configuration runs, each containing all 43 requests. Across three runs per arm, \sys{} records 28.13 pooled tokens/s and five-token native MTP records 26.12, a 7.7\% difference on this corpus. Their individual run rates range from 27.40 to 28.78 and from 26.04 to 26.23 tokens/s, respectively. Plain autoregressive decoding records 10.54 tokens/s in one run. The native MTP sweep uses one, three, five, or seven draft tokens; five-token MTP has the highest observed pooled rate among these settings.

\begin{table}[t]
\caption{Sampled replay of 43 recorded requests, capped at 1,024 output tokens. Rates pool matched token and time totals across all runs. Ranges describe individual run rates, not confidence intervals. SGLang receives the same input token IDs but uses a different stack and streaming endpoint.}
\label{tab:request-replay}
\centering\small
\begin{tabular}{@{}lrrr@{}}
\toprule
Configuration & Runs & Tokens/s & Run range \\
\midrule
vLLM autoregressive & 1 & 10.54 & --- \\
vLLM MTP, 1 token & 1 & 15.37 & --- \\
vLLM MTP, 3 tokens & 1 & 22.06 & --- \\
vLLM MTP, 5 tokens & 3 & 26.12 & 26.04--26.23 \\
vLLM MTP, 7 tokens & 1 & 25.72 & --- \\
\sys{} & 3 & 28.13 & 27.40--28.78 \\
\midrule
SGLang EAGLE, 7/8 & 3 & 29.69 & 28.91--30.90 \\
\bottomrule
\end{tabular}
\end{table}

For SGLang, the paired setting denotes speculative steps and draft-token capacity; EAGLE top-$k$ is one. Seven steps and eight draft tokens yield 29.69 pooled tokens/s over three runs with identical incoming token IDs. Attention, prefill settings, and endpoint timing still differ (Section~\ref{sec:limitations}). The native MTP sweep is exploratory and has unequal repetition counts.

Additional runs disable individual \sys{} optimizations (Table~\ref{tab:replay-variants}). Tables~\ref{tab:replay-variants}--\ref{tab:drafting-passes} predate the penalty-history correction of Section~\ref{sec:sampling-validation} and compare configurations under one sampler; the deployed rows in Tables~\ref{tab:request-replay} and~\ref{tab:disjoint-replay} are post-correction. The stock-attention variant completes all 43 requests in three runs and records 26.82 pooled tokens/s. Disabling fused draft selection or precomputed tap-operand preparation records 29.40 and 29.02 in one run each. Stochastic output lengths and limited replication prevent attributing these differences to the individual transformations. Phase and boot-allocation observations below characterize selected variants, without a complete per-optimization decomposition.

\begin{table}[t]
\caption{Exploratory replay variants. Each run contains all 43 requests. These measurements do not establish isolated optimization effects.}
\label{tab:replay-variants}
\centering\small
\setlength{\tabcolsep}{3pt}
\begin{tabular}{@{}lrrr@{}}
\toprule
Disabled optimization & Runs & Tokens/s & Run range \\
\midrule
Grouped split-K tree attention & 3 & 26.82 & 26.67--27.11 \\
Fused draft top-$k$ selection & 1 & 29.40 & --- \\
Precomputed tap-operand preparation & 1 & 29.02 & --- \\
\bottomrule
\end{tabular}
\end{table}

Phase timers add a descriptive view of these variants (Table~\ref{tab:optimization-spans}). Stock attention has a 14.6\,ms longer target span and a 15.1\,ms longer step than the two instrumented deployed repeats. Disabling fused selection increases the drafter span by 0.7\,ms; operand-preparation removal has no clear measured separation. Request inputs are shared, but generated trajectories, acceptance lengths, and actual KV capacities differ, so these are not isolated causal effects at identical computation.

\begin{table}[t]
\caption{Mean execution spans of exploratory replay variants, in milliseconds. Each phase pools its own duration and timer count. Deployed uses two instrumented runs; stock attention three; other variants one.}
\label{tab:optimization-spans}
\centering\small
\setlength{\tabcolsep}{3pt}
\begin{tabular}{@{}lrrrr@{}}
\toprule
Configuration & Target & Draft & Dispatch & Step \\
\midrule
Deployed & 112.14 & 49.84 & 20.81 & 189.35 \\
Stock attention & 126.78 & 49.94 & 21.13 & 204.44 \\
Fused selection off & 112.32 & 50.57 & 21.08 & 190.38 \\
Operand preparation off & 111.85 & 49.85 & 20.75 & 188.97 \\
\bottomrule
\end{tabular}
\end{table}

Boot logs report 20.54\,GiB of model-loading allocation in every variant. Reported CUDA-graph pools are 1.52--1.59\,GiB for the deployed runs, 0.42--0.50\,GiB for stock attention, 1.97\,GiB without fused selection, and 1.47\,GiB without prepared operands. Deployed KV capacity is 203,776 tokens versus 197,632--198,656 for stock attention. These allocation reports do not measure peak serving memory, and the unequal capacities further limit attribution.

Output-side counters pooled over the repeated vLLM runs report 4.35 accepted drafts per speculative event for \sys{} and 3.17 for five-token MTP. These counters describe emitted drafts, excluding the correction or bonus token, and do not identify the committed path. SGLang reports only windowed acceptance averages, so no corresponding cumulative acceptance comparison is reported.

\paragraph{Disjoint request corpus}
Table~\ref{tab:disjoint-replay} reports a separately frozen 43-request corpus from four additional Astropy task conversations. Across three runs per vLLM arm, \sys{} records 30.76 pooled tokens/s versus native MTP-5's 27.52, an 11.8\% difference. Every individual tree rate exceeds every MTP-5 rate in these runs. This second corpus supports the direction of the observed replay difference, without establishing statistical significance, equal output distributions, or task-completion improvement.

\begin{table}[t]
\caption{Disjoint 43-request replay. Rates pool matched totals; ranges are individual run rates. Request messages are shared, but SGLang uses a different chat serialization.}
\label{tab:disjoint-replay}
\centering\small
\begin{tabular}{@{}lrrr@{}}
\toprule
Configuration & Runs & Tokens/s & Run range \\
\midrule
\sys{} & 3 & 30.76 & 30.38--31.14 \\
vLLM MTP, 5 tokens & 3 & 27.52 & 27.41--27.71 \\
SGLang EAGLE, 7/8 & 1 & 30.24 & --- \\
\bottomrule
\end{tabular}
\end{table}

SGLang records 30.24 tokens/s in its single run. Limited cross-stack replication, different output lengths, and pooled duration weights preclude a superiority or equivalence claim. The small pooled tree--SGLang difference changes sign between the two corpora. SGLang's serialized prompt lengths exceed vLLM's by 33--156 tokens here, unlike the exact-token-ID comparison in Table~\ref{tab:request-replay}.

\paragraph{Drafting-pass sensitivity}
The deployed tree's maximum depth is eleven, but neural drafting uses only five MTP passes; suffix lookup supplies the deeper tail. We therefore measure three-pass drafting and a suffix-dependent three/five-pass policy while retaining 32 physical verification rows (Table~\ref{tab:drafting-passes}). These are drafting-policy variants, not measurements of newly constructed 16-node or shallower tree topologies.

\begin{table}[t]
\caption{Exploratory drafting-pass variants on the first 43-request corpus. The deployed row pools two runs; each variant has one. GPU drafting and step wall durations use their own timer counts.}
\label{tab:drafting-passes}
\centering\small
\setlength{\tabcolsep}{3pt}
\begin{tabular}{@{}lrrrr@{}}
\toprule
Policy & Tokens/s & \shortstack{Accepted\\drafts/event} & \shortstack{Draft\\ms} & \shortstack{Step\\ms} \\
\midrule
Five passes & 28.69 & 4.45 & 49.84 & 189.35 \\
Three passes & 28.94 & 3.97 & 32.01 & 171.27 \\
Conditional three/five & 27.05 & 4.09 & 47.71 & 187.55 \\
\bottomrule
\end{tabular}
\end{table}

Three-pass drafting saves approximately 18\,ms per cycle but reduces accepted drafts per event by about 11\%. Its 0.9\% pooled-rate difference from the deployed runs is small relative to the observed variation of the repeated configuration; no reliable gain is established. The conditional policy has a lower pooled rate. The five-pass configuration is retained. An initial shortened-pass implementation duplicated principal tokens into inactive sibling leaves, producing misleadingly low acceptance; the reported variants use the repaired implementation. Those earlier measurements and the correction remain in the supplement.

\subsection{Acceptance and execution costs}
The task study emits an average of 4.21 accepted drafts per \sys{} speculative event. This output-side counter excludes the correction or bonus token and does not identify the branch published into persistent state. Physical draft slots include padding and are distinct from logical candidates.

\begin{table}[t]
\caption{\sys{} acceptance and mean execution spans. Dispatch includes acceptance, path handling, and state commitment. GPU timers divide each duration sum by its own span count; wall time covers the decoding step. Disjoint replay pools three runs.}
\label{tab:recovered-telemetry}
\centering\small
\begin{tabular}{@{}lrr@{}}
\toprule
Metric & Task study & Disjoint replay \\
\midrule
Accepted drafts per event & 4.21 & 4.73 \\
Target GPU span (ms) & 115.43 & 109.84 \\
Drafter GPU span (ms) & 55.64 & 48.17 \\
Dispatch GPU span (ms) & 20.27 & 20.99 \\
Step wall time (ms) & 198.88 & 185.16 \\
\bottomrule
\end{tabular}
\end{table}

Table~\ref{tab:recovered-telemetry} characterizes the complete \sys{} cycle on these workloads. Dispatch combines sampling and native replay; the GPU spans do not form an isolated critical-path decomposition. Acceptance and phase durations vary with the request corpus, so higher replay acceptance is not evidence of better task quality.

\subsection{Numerical and candidate-selection validation}
\label{sec:current-validation}
\begin{table}[t]
\centering
\caption{GDN component checks on GB10. Bitwise columns compare with the native chain; the paired rule compares both implementations with a float64 reference. Counts are per block and process.}
\label{tab:gdn-component}
\small
\setlength{\tabcolsep}{3pt}
\begin{tabular}{@{}lrrr@{}}
\toprule
Operand sets & \shortstack{BF16 output\\bitwise} & \shortstack{FP32 state\\bitwise} & \shortstack{Paired rule\\passed} \\
\midrule
Calibration & 3,004/3,072 & 2,688/2,688 & 5,760/5,760 \\
Held-out & 3,018/3,072 & 2,688/2,688 & 5,760/5,760 \\
\bottomrule
\end{tabular}
\end{table}

We test GDN verification and accepted recurrent-state publication on two calibration and two held-out synthetic operand sets. Each set supplies 48 recurrent instances with the evaluated tree geometry: 32 physical output rows, including padding, and 28 accepted publication paths. Each block runs twice in each of two fresh processes. These repetitions check reproducibility; they do not increase the number of distinct inputs.

For each head, we compare both implementations with a float64 recurrence. We require both root-mean-square and maximum absolute errors to satisfy
\begin{equation}
 E_{\mathrm{tree}} \leq 1.10 E_{\mathrm{native}} + 2^{-24} M_{64} + 2^{-149},
\end{equation}
where $M_{64}$ is the corresponding root-mean-square or maximum magnitude of the reference. Table~\ref{tab:gdn-component} reports all enumerated cases passing in both processes. All published recurrent states are bitwise equal to the native chain, while 68 calibration and 54 held-out output tensors differ in BF16 rounding. Native and candidate results reproduce bitwise across repeats and processes.

Calibration controls detect sibling substitution, swapped state rings, and stale publication metadata. Poisoning an off-path branch leaves the tested selected-path state unchanged, and a zero-replay control is rejected. These results support the tested GDN scan and recurrent publication mechanism.

The fused full-vocabulary selection test compares the two output tensors against the reference argmax and top-k operations. Across 1,368 input cases and five block-count settings (6,840 configurations), it records zero byte mismatches; planted negative controls fire. A separate captured four-level selection graph completes 24 replays with zero mismatches. The reference selection outputs match exactly on these inputs and graph replays.

Attention tests use the same kernel build as the ten-task deployment. They show bit-identical repeated output and log-sum-exp values within each of sixteen determinism cases, with matching digests across two processes. On synthetic query/key/value tensors drawn at scales measured from model operands, 93.31\% of output elements agree with the reference kernel within two ULP, the maximum absolute output difference is $2^{-8}$, and maximum log-sum-exp disagreement is four ULP. All nine checks pass, including comparisons against a float64 dense reference and checks for non-finite disagreements. Split-K changes the reduction order, so its bounded error accompanies the repeatability result.

Runtime traces confirm execution of the specialized kernel in all sixteen attention layers, fused selection, device acceptance, and captured state publication during the workload runs.

\subsection{Full-model continuation}
\label{sec:fullmodel-validation}
The full-model comparison evaluates six fresh prefixes of approximately 23,000--46,000 tokens. For each prefix, three schedules replay harvested natural accepted paths over seven cycles, including a final root-only cycle that consumes the last pending token. The 18 schedules therefore cover 126 continuation cycles. Two schedules import a common starting state; the third resumes from each engine's own prefix-cache state. Every candidate forward is checked against all 32 recorded tree-input rows. These are replayed path schedules, not a test of the live sampler's complete output distribution. Appendix~\ref{app:fullmodel-protocol} specifies the five serving processes and controls.

For state or logit surface $s$ at cycle $c$, the frozen reference envelope is
\begin{equation}
 R_s(c)=\max\{F_s,d_s(B,A;c),d_s(V,A;c),d_s(P,A;c)\},
\end{equation}
where $A$ is the sequential reference, $B$ a separate process, $V$ a native cache-capacity control, and $P$ a native packed-kernel variant. State distance is the maximum per-layer relative $L_2$ error, with floor $F_s=0.02$; next-token KL divergence uses $F_s=0.15$ nats. The candidate must satisfy $d_s\leq2R_s$ on every surface and cycle. Constants were calibrated on an earlier collection and frozen before these fresh prefixes were evaluated.

All 126 cycles satisfy the numerical rule (Table~\ref{tab:fullmodel-agreement}), and all ten injected faults are detected at their mutation cycles. The candidate greedy token matches $A$ on 121/126 cycles; the five disagreements occur where the reference's top two logits tie exactly and satisfy the frozen logit-margin allowance. The separate-process, cache-capacity, and packed-kernel controls agree on 125, 125, and 123 cycles, respectively. Thus the result establishes agreement under the stated bounds, not bitwise model equality or strict greedy-output equality.

\begin{table}[t]
\caption{Full-model continuation over 126 cycles. Maximum candidate error and candidate-to-native-envelope ratio may occur at different cycles. Every ratio is below the frozen limit of 2. State errors are relative $L_2$; KL is in nats.}
\label{tab:fullmodel-agreement}
\centering\small
\setlength{\tabcolsep}{4pt}
\begin{tabular}{@{}lrr@{}}
\toprule
Surface & Maximum error & Maximum ratio \\
\midrule
GDN recurrent state & 0.422 & 1.09 \\
Convolution history & 0.350 & 1.11 \\
KV, newly committed positions & 0.373 & 1.34 \\
KV, earlier positions & 0.248 & 1.07 \\
Next-token KL divergence & 0.095 & 0.63 \\
\bottomrule
\end{tabular}
\end{table}

The ten faults comprise five types on two prefixes: omitted convolution gathering, omitted recurrent replay, incorrect KV remapping, stale root input, and sibling-state publication. Every cache-resumed case starts with at least 91.9\% of its prefix cached, across all five arms (30 observations). One candidate repeat is bitwise identical over its seven cycles. This covers prefix restoration on the tested inputs, rather than the full cache-admission and eviction lifecycle.

This collection uses the corrected sampler of Section~\ref{sec:sampling-validation}; an earlier held-out collection with the pre-correction sampler satisfied the same frozen rule and, with the inconclusive calibration collection, is preserved in the supplement. MTP drafter state and batch sizes above one are outside this comparison.

\subsection{Sampling scope and penalty histories}
\label{sec:sampling-validation}
\paragraph{Scope of exact sampling}
The sampling rule is exact for the intended target distribution only when the supplied probabilities are correct. History-dependent processing (presence, frequency, and repetition penalties) must therefore use each verification row's own ancestry: a target row includes the path through its parent, and a leaf continuation row includes the leaf itself. The deployed processor builds these per-path histories from the tree descriptor's parent table; the stock chain combiner, which gives row $j$ the flattened history of the first $j$ drafts, is retained for chain requests only. Two differences from native single-row decoding remain: the tree path ignores min-$p$, and its top-$k$ implementation retains exactly $k$ tokens, whereas native decoding retains boundary ties. The numerical continuation result does not prove joint output-distribution equality.

\paragraph{Captured-row and sampler audit}
A passive capture of 40 tree steps on four requests, taken with the deployed sampler at presence penalty 1.0, contains 2,480 processed target and continuation rows, including padded slots. All 2,480 carry the per-path history; 82 coincide with the flattened-chain history where the two agree. An earlier capture before this correction found all 2,480 rows flattened, with node-level total variation above 0.01 on approximately 22\% of deeper decisions and maximum 0.421; that defect, its repair, and the superseded measurements are in the supplement. Every measurement here uses the corrected sampler unless stated otherwise.

The deployed target-only walk consumes target and continuation logits without a separate draft-probability tensor. Its CPU floating-point oracle is sampled 200,000 times on each of six captured steps, for 1.2 million walks. Tests compare emitted-token counts conditional on visiting a node with the supplied probability row, using an approximate chi-square $G$-test and a Bonferroni-adjusted per-step threshold of $10^{-3}$. Nodes require at least 2,000 visits; rare expected counts are pooled. No departure is detected in 48 node tests against the supplied rows. Against autoregressive rows reconstructed with the native tie-keeping top-$k$, one of 48 node tests rejects; over all 1,120 node-steps the reconstructed total variation is at most 0.028, and the three nodes above 0.01, including the rejected one, are exact ties at the top-$k$ boundary. On these rows, no departure from the autoregressive target is found beyond the stated tie convention. All five planted fault types are detected in at least one step (Table~\ref{tab:sampling-faults}). Non-rejection supports this finite check; it does not prove universal exactness.

\begin{table}[t]
\caption{Sampler-audit fault sensitivity on six captured steps. Each fault type is detected at least once; detection depends on which nodes the walk visits.}
\label{tab:sampling-faults}
\centering\small
\begin{tabular}{@{}lr@{}}
\toprule
Injected fault & Steps detected \\
\midrule
Temperature applied twice & 5/6 \\
Target rows shifted & 6/6 \\
Fixed residual draw & 4/6 \\
Unconditional acceptance & 3/6 \\
Correlated random draws & 1/6 \\
\bottomrule
\end{tabular}
\end{table}

An earlier ten-boot, 8,000-continuation output diagnostic failed its frozen temperature-control sensitivity requirement and remains uninformative. The tree sampler also seeds one process-wide generator with zero, so boots with identical request order reuse random sequences. They cannot be treated as independent sampling replicates. The captured-row audit above is informative where that output test was not.

\subsection{Recurrent agreement under distinct references}
\label{sec:m1-qualification}
\begin{table}[t]
\centering
\caption{Recurrent agreement on GB10 under distinct sequential references. Counts give cells outside the paired error bounds among 64,512 output cells and 6,912 state cells per configuration, with identical counts in both repeats. The two reference groups are not a common accuracy ranking.}
\label{tab:m1-qualification}
\small
\setlength{\tabcolsep}{3pt}
\begin{tabular}{@{}lrr@{}}
\toprule
Configuration & Output cells & State cells \\
\midrule
\multicolumn{3}{@{}l}{\textit{Native GPU recurrence reference}} \\
\sys{} & 0 & 0 \\
\midrule
\multicolumn{3}{@{}l}{\textit{Sequential software references}} \\
Weaver author-default & 53,745 & 1,154 \\
Weaver (FP32 preparation) & 57,707 & 388 \\
TreeWY author-default & 37,528 & 6,912 \\
\bottomrule
\end{tabular}
\end{table}

We compare recurrent verification and accepted-state publication using held-out synthetic operands spanning 48 recurrent instances and 48 value heads. Verification covers the tree's 28 active nodes. Three accepted-path publications produce a history of 1, 6, and 11 updates, including Weaver's accepted replay and TreeWY's final deferred commit. Each configuration is evaluated twice in one process, with a fresh initial state for each cycle. A numerical cell fixes the recurrent instance, output node or publication, and value head; errors are reduced over the remaining vector or matrix entries.

Each cell must satisfy both root-mean-square and maximum absolute error bounds,
\begin{equation}
 E_{\mathrm{method}} \leq 1.10 E_{\mathrm{seq}} + 2^{-24}M_{64}+2^{-149},
\end{equation}
where errors are measured against the specified float64 recurrence and $M_{64}$ is its corresponding root-mean-square or maximum magnitude. For \sys{}, the sequential comparator is the native GPU recurrence. Weaver and TreeWY use sequential software comparators that do not emulate every tensor-core rounding operation.

TreeWY retains the author implementation's normalization floor of $10^{-12}$ and BF16 dot products. Weaver uses TF32 verification, FP32 elementwise accepted replay, and an additive normalization constant of $10^{-6}$. We report both its author-default preparation and a variant with FP32 preparation. The two Weaver variants retain their respective preprocessing conventions.

Table~\ref{tab:m1-qualification} reports the cells outside each implementation's paired bounds. All configurations produce finite tensors and reproduce bitwise between repeats. TreeWY's BF16 compact commit and its sequential comparator's FP32 outer-product updates, for example, are different floating-point computations. The normalization and arithmetic conventions are part of the numerical comparison and must be interpreted alongside its reference choice (Section~\ref{sec:limitations}).

\subsection{Recurrent verification and publication costs}
\label{sec:mechanism-costs}
We measure verification and accepted-state publication on shared synthetic operands for 48 GDN instances with the model's head dimensions, using the 28 active nodes of the deployed 32-row tree. Each measurement has 20 warmup iterations and 100 timed iterations. CUDA events bracket verification and publication; a 256-MiB buffer is written outside each timed iteration. Graph mode captures the verifier while retaining each method's publication wrapper; eager mode invokes the verifier directly. Accepted paths cycle through all 28 possible frontiers. Durable state is reset before each measurement and evolves across iterations. These measurements exclude attention, convolution, sampling, model projections, and interleaved execution of the remaining model layers.

\begin{table}[t]
\caption{GDN component latency in milliseconds on GB10. Verify, publish, and total columns give graph-mode medians; eager gives the total median without verifier graph replay. TreeWY fuses publication of the previous frontier with the current verification; its final flush is reported separately. These are timing observations under distinct numerical policies.}
\label{tab:mechanism-times}
\centering\small
\setlength{\tabcolsep}{3pt}
\begin{tabular}{@{}lrrrr@{}}
\toprule
Method & Verify & Publish & Total & Eager \\
\midrule
\sys{} & 12.90 & 4.49 & 17.69 & 17.58 \\
Native path baseline & 29.71 & 2.70 & 32.41 & 33.73 \\
PyTorch node baseline & 52.32 & 2.70 & 55.02 & 81.04 \\
TreeWY author-default & \multicolumn{2}{c}{fused} & 11.01 & 11.71 \\
Weaver author-default & 2.07 & 1.38 & 3.45 & 5.88 \\
Weaver (FP32 preparation) & 4.05 & 1.40 & 5.45 & 9.13 \\
\bottomrule
\end{tabular}
\end{table}

The native path baseline uses native recurrent updates with the same path cover but stores every node's full state. The PyTorch baseline instead updates nodes sequentially in topological order. Both publish by copying the selected leaf into durable state. \sys{} records a 17.69-ms graph-mode total, with a 10th--90th percentile interval of 17.18--17.84 ms, versus 32.41 ms (32.36--32.67) for native paths. Total medians are measured directly; they need not equal sums of region medians.

TreeWY's steady step takes 11.01 ms. A finite sequence also requires its initial verification without a pending commit and a final deferred publication; the latter takes 10.51 ms in this implementation. Weaver includes accepted-path replay in its total. Its graph-mode topology rebuild takes approximately 0.02 ms separately and is excluded from the table because the topology is reused. These boundaries must be retained in a complete-cycle comparison.

\begin{table}[t]
\caption{Named persistent buffer allocations and numerical error for the GDN measurements. Output and state columns are maximum absolute errors against the float64 recurrence; state uses the deepest accepted path. Buffer totals exclude outputs, separate page pools, graph pools, and temporary per-call allocations; the naive state banks include reserved null and durable slots. These are not peak system memory or measured memory traffic.}
\label{tab:mechanism-memory-error}
\centering\small
\setlength{\tabcolsep}{3pt}
\begin{tabularx}{\columnwidth}{@{}Yrrr@{}}
\toprule
Method & Buffers (MiB) & Output error & State error \\
\midrule
\sys{} & 132.5 & $4.82\times 10^{-4}$ & $2.62\times 10^{-6}$ \\
Native path baseline & 5,040.0 & $4.82\times 10^{-4}$ & $2.62\times 10^{-6}$ \\
PyTorch node baseline & 5,040.0 & $4.82\times 10^{-4}$ & $6.51\times 10^{-7}$ \\
TreeWY author-default & 192.8 & $6.31\times 10^{-4}$ & $5.10\times 10^{-3}$ \\
Weaver author-default & 25.1 & $5.28\times 10^{-4}$ & $5.06\times 10^{-3}$ \\
Weaver (FP32 preparation) & 31.1 & $5.18\times 10^{-4}$ & $8.55\times 10^{-7}$ \\
\bottomrule
\end{tabularx}
\end{table}

Table~\ref{tab:mechanism-memory-error} sums the tensor allocations explicitly retained for verification and publication. \sys{}'s 96-MiB cut-state buffer is reused across layers; only five cut rows, totaling 15 MiB per layer, are logically exported. This agrees with the storage formula in Section~\ref{sec:kernel}, but does not measure physical memory transactions. Operand rings and publication staging contribute the remaining named buffers. The deployment's separately allocated 1.5-GiB recurrent page pool is excluded. The naive baselines retain a 34-slot bank (32 physical nodes, one null slot, and one durable slot) and a staging copy for publication. Thus allocation boundaries differ across implementations. Framework peak allocations and all raw timing samples are preserved in the supplement.

The errors in this experiment are descriptive, rather than an admission criterion. TreeWY and the two Weaver variants retain the preprocessing and precision differences described in Section~\ref{sec:m1-qualification}. Consequently, faster timings here do not establish equal-accuracy superiority, and they do not overturn the paired-bound failures in Table~\ref{tab:m1-qualification}. These isolated GDN measurements also do not predict a full-model decoding speedup.

\section{Discussion}
\label{sec:discussion}
The design couples tree geometry, physical layout, and continuation publication. Optimizing any one of these in isolation can move work elsewhere or invalidate another stage's mapping. Transient cut states, graph buffers, and native persistent rows serve different purposes: their separation makes the storage and recomputation tradeoff explicit. Path scans retain sequential arithmetic, while native replay pays accepted-path update work to return persistent state to the serving representation.

For coding agents, decoding rate and task completion measure different effects. Additional tokens can reflect more reasoning or a longer tool loop, and tool execution contributes to agent duration without entering the decode denominator. Recorded-request replay fixes incoming messages but does not execute the newly generated tools or subsequent agent decisions. The measured 7.7\% and 11.8\% rate advantages over five-token native MTP therefore apply to the two request corpora and serving configuration. Against SGLang's EAGLE configuration, whose top-$k$ of one makes its seven-step draft a chain, the pooled difference is within 5\% and changes sign between corpora: \sys{} reaches a chain stack's rate while verifying 27-candidate trees, so that comparison measures the cost of hosting tree verification, not a gain from it. Tree drafters, which chain verification cannot host, are not evaluated here. They do not transfer to a task-level speed claim: in the live agent study, \sys{} generates more tokens, takes longer in aggregate, and resolves five tasks compared with MTP's six. That gap is not a sampling effect: on fixed replay prompts the corrected tree arm and MTP-5 produce indistinguishable output lengths (18,916 versus 18,504 mean tokens on the first corpus, 15,270 versus 15,453 on the disjoint corpus; tree longer on 23 and 19 of 43 requests), so the live-agent difference arises from diverging trajectories. Task outcomes and GPU spans describe complementary parts of the serving/application boundary.

\section{Limitations}
\label{sec:limitations}
\paragraph{Numerical scope}
The component results cover recurrent verification and publication, attention error, and candidate selection. The held-out full-model comparison covers 126 replayed continuation cycles with sensitive fault controls. It permits native-relative numerical error and margin-bounded greedy disagreements, rather than establishing bitwise or strict greedy equality. Harvested paths do not validate the live acceptance loop or MTP drafter state, and 18 schedules do not exhaust every path combination. Prefix restoration is covered on the tested prompts; cache admission and eviction, row reuse, batch sizes above one, and complete distribution preservation remain outside that result. The sampler audit covers captured rows and finite Monte-Carlo walks on six steps with presence penalty 1.0; the tree path's exact-$k$ top-$k$ convention and ignored min-$p$ differ from single-row decoding, and finite non-rejection cannot prove universal exactness. The paired recurrent tests use different sequential comparators and preprocessing conventions, so they do not establish a common accuracy ranking. Their isolated timings omit the rest of the hybrid model. Matching starting states, tokens, and positions does not guarantee identical intermediate hidden tensors or identify the cause of native variation.

\paragraph{Performance and task scope}
The task study contains one sampled attempt per task and arm, ten selected Astropy tasks, and no plain-decoding agent control. It provides neither a benchmark-wide quality estimate nor an isolated task-completion speedup: the arms execute different request and tool trajectories. Both replay corpora have three repetitions per principal vLLM arm; the disjoint corpus has one SGLang run and is drawn from four additional task conversations within the same project. Repeated tree boots reuse a process-seeded random sequence and are not independent sampling replicates. This limited replication does not establish significance, SGLang parity, or broad workload generalization. Sampling changes output lengths, and the response cap truncates some requests. Stream buffering and parsing enter the client-side duration; subtracting one token approximates the first output interval when a speculative chunk contains several tokens. Exact input IDs remove the serialization discrepancy only in the first corpus's SGLang comparison. The disjoint replay and live agent study use separate chat templates, attention and prefill implementations. Optimization variants have limited repetitions and do not isolate every transformation. Equal memory-fraction settings do not establish equal KV-block capacity. GPU instrumentation overhead has not been measured. Boot allocation logs do not establish peak serving memory. Named scratch allocations exclude other resident buffers and temporary work; logical state-export bytes are not measurements of memory traffic.

\section{Conclusion}
\sys{} combines path-parallel recurrent scans with native replay and a shared mapping for convolution, attention, acceptance, and drafter continuation. Sampled replay on GB10 yields 28.13 versus 26.12 pooled tokens/s for five-token native MTP, and 30.76 versus 27.52 on a disjoint corpus, with three runs per arm. It verifies 27-candidate trees within 5\% of a tuned chain-speculation stack, so tree verification is available on a hybrid model at chain-stack cost; tree drafters are the untested next step. The 30-attempt task study records 26.12 versus 27.17 tokens/s and five versus six resolved tasks. Full-model continuation satisfies frozen native-relative numerical bounds over 126 cycles and detects all ten injected faults. The deployed sampler applies per-path penalty histories; a captured-row audit and a Monte-Carlo test of the acceptance walk find no departure from the autoregressive target beyond the top-$k$ tie convention, and identical replay prompts yield indistinguishable output lengths under the tree and chain arms. The results characterize bounded numerical agreement, sampler agreement on captured rows, and replay rates of the implemented system, without claiming task-completion speedup.

\appendices
\section{Serving Configuration and Measurement}
\label{app:audit}
\label{app:settings}
The implementation artifact~\cite{lumo2026archive} documents the serving route and component studies. This revision's source and data supplement preserves the additional task study, recorded-request replays, continuation diagnostics, and recurrent cost measurements. Input manifests, run selections, pooled-rate reductions, allocation accounting, raw timing samples, and failed or invalid observations remain separate from the manuscript's scientific claims.
\begin{table}[!htbp]
\caption{Serving settings for the three-arm task study. Native MTP and \sys{} share the vLLM build. Exact versions and launch settings accompany the supplement.}
\label{tab:serving-config}
\centering\footnotesize
\renewcommand{\arraystretch}{1.12}
\begin{tabularx}{\columnwidth}{@{}>{\raggedright\arraybackslash}p{0.22\columnwidth}Y@{}}
\toprule
Setting & Configuration \\
\midrule
Target & Qwen3.8-27B; mixed NVFP4/FP8 weights, BF16 execution \\
Drafter & Same-checkpoint one-layer BF16 MTP in all arms; no separate EAGLE draft weights \\
\sys{} proposals & 27 drafts plus root, padded to 32 rows; maximum depth 11; five MTP passes and suffix extension \\
Native proposals & MTP: five drafts; SGLang: seven steps, top-$k$ one, capacity eight drafts \\
Attention & vLLM: forked FlashAttention-2; \sys{} adds grouping 2 and split-K 4; SGLang: FlashInfer \\
Scheduling & One active sequence; 4,096 prefill tokens for vLLM, 8,192 for SGLang \\
Context / memory & 131,072 tokens; memory fraction 0.70; actual KV capacities not equalized \\
Agent / budget & Qwen Code 0.19.4; one task at a time; 9,000 seconds \\
Response cap & 24,000 tokens per request \\
Sampling & Temperature 0.6; top-$p$ 0.95; top-$k$ 20; min-$p$ 0; presence penalty 1.0 \\
\bottomrule
\end{tabularx}
\end{table}

All tree runs use presence penalty 1.0. Tables~\ref{tab:agent-study}, \ref{tab:request-replay}, \ref{tab:disjoint-replay}, \ref{tab:recovered-telemetry}, \ref{tab:fullmodel-agreement}, and~\ref{tab:swe-tasks} use per-path penalty histories (Section~\ref{sec:sampling-validation}); Tables~\ref{tab:replay-variants}--\ref{tab:drafting-passes} predate that correction. The exact-$k$ top-$k$ convention and ignored min-$p$ remain the differences from single-row decoding.

Workload rates use Eq.~\ref{eq:pooled-decode}; token counts and latency sums cover the same completed requests. Metric counters are checked for continuity, resets, and outstanding requests at task boundaries. First-output latency, tools, and evaluator time remain distinct from decode duration. GPU phase times divide each timer's duration sum by its own span count. Output-side acceptance counters describe emitted draft lengths and do not identify the committed tree path.

For target probabilities $p$ and proposal probabilities $q$, ordinary rejection sampling combines accepted mass $\min(p,q)$ with residual mass proportional to $(p-q)_+$. This gives the target marginal for the probabilities actually supplied~\cite{leviathan2023fast,chen2023sampling}. Serving additionally requires that subsequent forwards consume the state associated with the accepted path and that logits processing constructs that path's target distribution. The deployed penalty processing builds each verification row's history from its own root-to-node path (Section~\ref{sec:sampling-validation}); the remaining differences from single-row decoding are the top-$k$ tie convention and the ignored min-$p$. Artifact version~2 includes the corrected sampler (commit \texttt{9902ccca2}), the captured rows, and all post-correction records; version~1 predates the correction.

\section{Additional Workload Results}
\label{app:workload}
Table~\ref{tab:swe-tasks} retains every attempt in the three-arm task study; the \sys{} column is the attempt with the corrected sampler; the earlier pre-correction attempt is preserved in the supplement. Agent duration includes model service and tool execution, excluding server startup and subsequent evaluation. Empty-patch outcomes are distinguished from patches that reach the test evaluator.

\begin{table}[!htbp]
\caption{Task-level outcomes. IDs have prefix \texttt{astropy\_\_astropy-}. Each cell gives rounded agent minutes and outcome: R, resolved; F, failed tests; E, empty patch.}
\label{tab:swe-tasks}
\centering\small
\begin{tabular}{@{}lccc@{}}
\toprule
Task suffix & \sys{} & MTP-5 & SGLang \\
\midrule
13977 & 11 / E & 10 / F & 21 / F \\
14096 & 18 / R & 20 / E & 10 / E \\
14182 & 24 / R & 6 / E & 1 / E \\
14309 & 11 / R & 2 / R & 2 / R \\
14365 & 10 / F & 2 / R & 3 / R \\
14369 & 25 / F & 18 / R & 16 / F \\
14508 & 38 / R & 7 / R & 7 / R \\
14539 & 6 / E & 3 / R & 17 / R \\
14598 & 37 / F & 27 / F & 20 / F \\
14995 & 11 / R & 5 / R & 5 / R \\
\bottomrule
\end{tabular}
\end{table}

\section{Full-Model Comparison Protocol}
\label{app:fullmodel-protocol}
The continuation comparison uses five fresh serving processes: sequential reference $A$, an independent process $B$, a native cache-capacity control $V$, a native packed-GDN-kernel variant $P$, and \sys{}. The reference uses a 40\,GiB KV allocation; $V$ uses the candidate's observed allocation. Candidate capacity follows its deployed memory-fraction policy. All arms consume matching forced tokens and positions and retain their own published state across subsequent cycles.

Six prefixes from four additional Astropy task conversations are disjoint from the earlier continuation and first replay corpora. A separate harvest records natural tree inputs and accepted paths. Each prefix supplies three seven-cycle schedules: a cold salted prefix with a common starting-state import, an unsalted repeat that primes the cache with the same import, and a cache-resumed case with no import. The last cycle accepts only the root to consume the final pending token; next-forward logits also observe the subsequent pending token. The root-only cycle is included in the 126-cycle total. Candidate-input checks compare every one of the 32 verification rows with the recorded tree.

The frozen numerical rule uses the largest same-cycle native-control distance, floored at 0.02 for state relative $L_2$ and 0.15 nats for next-token KL. Candidate distance must not exceed twice this envelope. A greedy mismatch is permitted only when $A$'s top-two logit margin is within $\max(0.25,2\max_{B,V,P}\|\Delta\mathrm{logit}\|_\infty)$ or the native controls are categorically unstable outside that band; at most 10\% of cells may be native-unstable. Input parity, complete sealed observations, starting-state read-back, and detection of every injected fault are prerequisites. These allowances make the verdict a numerical agreement test, rather than exact token or logit equality.

Five fault types run on two prefixes each at a selected natural cycle: omitted convolution gathering, omitted recurrent replay, incorrect KV remapping, stale root input, and sibling-state publication. Each must exceed the same base case's bound on its designated surface; sibling publication may exceed any of its three targeted state surfaces. Faults are reported separately from the 126 clean cycles. Prefix-cache coverage requires at least 90\% cached tokens in every cache-resumed observation. One clean candidate case is repeated in the same process. Earlier invalid forcing and insensitive comparison records remain in the supplement with their original dispositions.

\FloatBarrier
\label{ReferencesStart}
\bibliographystyle{IEEEtran}
\bibliography{ref}

\begin{thebibliography}{10}
\providecommand{\url}[1]{#1}
\csname url@samestyle\endcsname
\providecommand{\newblock}{\relax}
\providecommand{\bibinfo}[2]{#2}
\providecommand{\BIBentrySTDinterwordspacing}{\spaceskip=0pt\relax}
\providecommand{\BIBentryALTinterwordstretchfactor}{4}
\providecommand{\BIBentryALTinterwordspacing}{\spaceskip=\fontdimen2\font plus
\BIBentryALTinterwordstretchfactor\fontdimen3\font minus
  \fontdimen4\font\relax}
\providecommand{\BIBforeignlanguage}[2]{{%
\expandafter\ifx\csname l@#1\endcsname\relax
\typeout{** WARNING: IEEEtran.bst: No hyphenation pattern has been}%
\typeout{** loaded for the language `#1'. Using the pattern for}%
\typeout{** the default language instead.}%
\else
\language=\csname l@#1\endcsname
\fi
#2}}
\providecommand{\BIBdecl}{\relax}
\BIBdecl

\bibitem{leviathan2023fast}
\BIBentryALTinterwordspacing
Y.~Leviathan, M.~Kalman, and Y.~Matias, ``Fast inference from transformers via
  speculative decoding,'' in \emph{Proceedings of the 40th International
  Conference on Machine Learning}, ser. Proceedings of Machine Learning
  Research, vol. 202.\hskip 1em plus 0.5em minus 0.4em\relax PMLR, 2023, pp.
  19\,274--19\,286. [Online]. Available:
  \url{https://proceedings.mlr.press/v202/leviathan23a.html}
\BIBentrySTDinterwordspacing

\bibitem{chen2023sampling}
\BIBentryALTinterwordspacing
C.~Chen, S.~Borgeaud, G.~Irving, J.-B. Lespiau, L.~Sifre, and J.~Jumper,
  ``{Accelerating Large Language Model Decoding with Speculative Sampling},''
  2023. [Online]. Available: \url{https://arxiv.org/abs/2302.01318}
\BIBentrySTDinterwordspacing

\bibitem{miao2023specinfer}
\BIBentryALTinterwordspacing
X.~Miao, G.~Oliaro, Z.~Zhang, X.~Cheng, Z.~Wang, Z.~Zhang, R.~Y.~Y. Wong,
  A.~Zhu, L.~Yang, X.~Shi, C.~Shi, Z.~Chen, D.~Arfeen, R.~Abhyankar, and
  Z.~Jia, ``{SpecInfer}: Accelerating large language model serving with
  tree-based speculative inference and verification,'' in \emph{Proceedings of
  the 29th ACM International Conference on Architectural Support for
  Programming Languages and Operating Systems, Volume 3}.\hskip 1em plus 0.5em
  minus 0.4em\relax ACM, 2024, pp. 932--949. [Online]. Available:
  \url{https://doi.org/10.1145/3620666.3651335}
\BIBentrySTDinterwordspacing

\bibitem{yang2024gated}
\BIBentryALTinterwordspacing
S.~Yang, J.~Kautz, and A.~Hatamizadeh, ``Gated delta networks: Improving
  {Mamba2} with delta rule,'' in \emph{The Thirteenth International Conference
  on Learning Representations}, 2025. [Online]. Available:
  \url{https://openreview.net/forum?id=r8H7xhYPwz}
\BIBentrySTDinterwordspacing

\bibitem{wu2025stree}
\BIBentryALTinterwordspacing
Y.~Wu, Z.~Qin, A.~Wong, and S.~Soatto, ``{STree}: Speculative tree decoding for
  hybrid state-space models,'' 2025. [Online]. Available:
  \url{https://arxiv.org/abs/2505.14969}
\BIBentrySTDinterwordspacing

\bibitem{wang2026specla}
\BIBentryALTinterwordspacing
Z.~Wang, X.~Han, Z.~Yang, F.~Liu, X.~Li, R.~Gu, S.~Zhong, and C.~Tian,
  ``{SpecLA: Efficient Speculative Decoding for Linear-Attention Models},''
  2026. [Online]. Available: \url{https://arxiv.org/abs/2607.16673v1}
\BIBentrySTDinterwordspacing

\bibitem{oda2026marginals}
\BIBentryALTinterwordspacing
Y.~Oda, R.~Mathieu, R.~Knyazhitskiy, and A.~Chakhvadze, ``{Trees from
  Marginals: Autoregressive drafting with factorized priors},'' 2026, technical
  report. [Online]. Available: \url{https://arxiv.org/abs/2607.06763}
\BIBentrySTDinterwordspacing

\bibitem{jimenez2024swe}
\BIBentryALTinterwordspacing
C.~E. Jimenez, J.~Yang, A.~Wettig, S.~Yao, K.~Pei, O.~Press, and K.~Narasimhan,
  ``{SWE-bench}: Can language models resolve real-world {GitHub} issues?'' in
  \emph{The Twelfth International Conference on Learning Representations},
  2024. [Online]. Available: \url{https://openreview.net/forum?id=VTF8yNQM66}
\BIBentrySTDinterwordspacing

\bibitem{chen2024sequoia}
\BIBentryALTinterwordspacing
Z.~Chen, A.~May, R.~Svirschevski, Y.~Huang, M.~Ryabinin, Z.~Jia, and B.~Chen,
  ``{Sequoia}: Scalable and robust speculative decoding,'' in \emph{Advances in
  Neural Information Processing Systems}, vol.~37, 2024, pp.
  129\,531--129\,563. [Online]. Available:
  \url{https://doi.org/10.52202/079017-4116}
\BIBentrySTDinterwordspacing

\bibitem{li2024eagle}
\BIBentryALTinterwordspacing
Y.~Li, F.~Wei, C.~Zhang, and H.~Zhang, ``{EAGLE-2}: Faster inference of
  language models with dynamic draft trees,'' in \emph{Proceedings of the 2024
  Conference on Empirical Methods in Natural Language Processing}.\hskip 1em
  plus 0.5em minus 0.4em\relax Association for Computational Linguistics, 2024,
  pp. 7421--7432. [Online]. Available:
  \url{https://aclanthology.org/2024.emnlp-main.422/}
\BIBentrySTDinterwordspacing

\bibitem{wang2024opt}
\BIBentryALTinterwordspacing
J.~Wang, Y.~Su, J.~Li, Q.~Xia, Z.~Ye, X.~Duan, Z.~Wang, and M.~Zhang,
  ``{OPT-Tree}: Speculative decoding with adaptive draft tree structure,''
  2025. [Online]. Available: \url{https://arxiv.org/abs/2406.17276}
\BIBentrySTDinterwordspacing

\bibitem{li2025eagle3}
Y.~Li, F.~Wei, C.~Zhang, and H.~Zhang, ``{EAGLE-3}: Scaling up inference
  acceleration of large language models via training-time test,'' in
  \emph{Advances in Neural Information Processing Systems}, vol.~38, 2025, pp.
  136\,737--136\,756.

\bibitem{chen2026dflash}
\BIBentryALTinterwordspacing
J.~Chen, Y.~Liang, and Z.~Liu, ``{DFlash: Block Diffusion for Flash Speculative
  Decoding},'' 2026. [Online]. Available:
  \url{https://arxiv.org/abs/2602.06036}
\BIBentrySTDinterwordspacing

\bibitem{kwon2023efficient}
\BIBentryALTinterwordspacing
W.~Kwon, Z.~Li, S.~Zhuang, Y.~Sheng, L.~Zheng, C.~H. Yu, J.~E. Gonzalez,
  H.~Zhang, and I.~Stoica, ``Efficient memory management for large language
  model serving with {PagedAttention},'' in \emph{Proceedings of the 29th
  Symposium on Operating Systems Principles}.\hskip 1em plus 0.5em minus
  0.4em\relax ACM, 2023, pp. 611--626. [Online]. Available:
  \url{https://doi.org/10.1145/3600006.3613165}
\BIBentrySTDinterwordspacing

\bibitem{zheng2024sglang}
\BIBentryALTinterwordspacing
L.~Zheng, L.~Yin, Z.~Xie, C.~Sun, J.~Huang, C.~H. Yu, S.~Cao, C.~Kozyrakis,
  I.~Stoica, J.~E. Gonzalez, C.~Barrett, and Y.~Sheng, ``{SGLang}: Efficient
  execution of structured language model programs,'' in \emph{Advances in
  Neural Information Processing Systems}, vol.~37, 2024, pp. 62\,557--62\,583.
  [Online]. Available: \url{https://doi.org/10.52202/079017-2000}
\BIBentrySTDinterwordspacing

\bibitem{pan2024marconi}
\BIBentryALTinterwordspacing
R.~Pan, Z.~Wang, Z.~Jia, C.~Karakus, L.~Zancato, T.~Dao, Y.~Wang, and
  R.~Netravali, ``{Marconi: Prefix Caching for the Era of Hybrid LLMs},'' 2025.
  [Online]. Available: \url{https://arxiv.org/abs/2411.19379}
\BIBentrySTDinterwordspacing

\bibitem{pytorch2025sglang}
{SGLang Team}, ``{Hybrid Models Meet SGLang: More than Full Attention},''
  \url{https://pytorch.org/blog/hybrid-models-meet-sglang-more-than-full-attention/},
  2025.

\bibitem{sglang2026qwen}
{SGLang Project}, ``{Qwen3-Next SGLang documentation},''
  \url{https://lmsysorg.mintlify.app/cookbook/autoregressive/Qwen/Qwen3-Next},
  2026.

\bibitem{dao2023flashdecoding}
T.~Dao, D.~Haziza, F.~Massa, and G.~Sizov, ``{Flash-Decoding for long-context
  inference},'' \url{https://pytorch.org/blog/flash-decoding/}, 2023, {October}
  13.

\bibitem{ye2025flashinfer}
Z.~Ye, L.~Chen, R.~Lai, W.~Lin, Y.~Zhang, S.~Wang, T.~Chen, B.~Kasikci,
  V.~Grover, A.~Krishnamurthy, and L.~Ceze, ``{FlashInfer}: Efficient and
  customizable attention engine for {LLM} inference serving,'' in
  \emph{Proceedings of Machine Learning and Systems}, vol.~7, 2025.

\bibitem{yao2024deft}
\BIBentryALTinterwordspacing
J.~Yao, K.~Chen, K.~Zhang, J.~You, B.~Yuan, Z.~Wang, and T.~Lin, ``{DeFT}:
  Decoding with flash tree-attention for efficient tree-structured {LLM}
  inference,'' in \emph{The Thirteenth International Conference on Learning
  Representations}, 2025. [Online]. Available:
  \url{https://openreview.net/forum?id=2c7pfOqu9k}
\BIBentrySTDinterwordspacing

\bibitem{pan2025fasttree}
Z.~Pan, Y.~Ding, Y.~Guan, Z.~Wang, Z.~Yu, X.~Tang, Y.~Wang, and Y.~Ding,
  ``{FastTree: Optimizing Attention Kernel and Runtime for Tree-Structured LLM
  Inference},'' in \emph{Proceedings of Machine Learning and Systems}, vol.~7,
  2025.

\bibitem{dao2023flashattention}
\BIBentryALTinterwordspacing
T.~Dao, ``{FlashAttention-2}: Faster attention with better parallelism and work
  partitioning,'' in \emph{The Twelfth International Conference on Learning
  Representations}, 2024. [Online]. Available:
  \url{https://openreview.net/forum?id=mZn2Xyh9Ec}
\BIBentrySTDinterwordspacing

\bibitem{yang2024delta}
\BIBentryALTinterwordspacing
S.~Yang, B.~Wang, Y.~Zhang, Y.~Shen, and Y.~Kim, ``Parallelizing linear
  transformers with the delta rule over sequence length,'' in \emph{Advances in
  Neural Information Processing Systems}, vol.~37, 2024, pp.
  115\,491--115\,522. [Online]. Available:
  \url{https://doi.org/10.52202/079017-3668}
\BIBentrySTDinterwordspacing

\bibitem{wu2024snakes}
\BIBentryALTinterwordspacing
Y.~Wu, Y.~Dukler, M.~Trager, A.~Achille, W.~Xia, and S.~Soatto, ``{Snakes and
  Ladders: Accelerating SSM Inference with Speculative Decoding},'' in
  \emph{Proceedings of the 4th NeurIPS Efficient Natural Language and Speech
  Processing Workshop}, vol. 262.\hskip 1em plus 0.5em minus 0.4em\relax PMLR,
  2024, pp. 292--304. [Online]. Available:
  \url{https://proceedings.mlr.press/v262/wu24a.html}
\BIBentrySTDinterwordspacing

\bibitem{zhong2025specmamba}
\BIBentryALTinterwordspacing
L.~Zhong, S.~Xu, H.~Wen, T.~Xie, Q.~Guo, Y.~Wang, and M.~Li, ``{SpecMamba}:
  Accelerating {Mamba} inference on {FPGA} with speculative decoding,'' 2025.
  [Online]. Available: \url{https://arxiv.org/abs/2509.19873}
\BIBentrySTDinterwordspacing

\bibitem{borobia2026component}
\BIBentryALTinterwordspacing
H.~Borobia, E.~Seguí-Mas, and G.~Tormo-Carbó, ``Component-aware
  self-speculative decoding in hybrid language models,'' 2026. [Online].
  Available: \url{https://arxiv.org/abs/2605.01106}
\BIBentrySTDinterwordspacing

\bibitem{wang2026bole}
\BIBentryALTinterwordspacing
L.~Wang, Y.~Su, X.~Wu, C.~You, Y.~Liu, Z.~Qiu, J.~Zhang, J.~Zheng, F.~Liu,
  J.~Zhang, C.~Tian, and C.~Huan, ``{Bole: Efficient Tree Speculation for
  Hybrid-Attention Language Models},'' 2026. [Online]. Available:
  \url{https://arxiv.org/abs/2608.01651}
\BIBentrySTDinterwordspacing

\bibitem{ghantasala2026treewy}
\BIBentryALTinterwordspacing
S.~M. Ghantasala, ``{TreeWY: Speculative Verification for Gated DeltaNet
  Hybrids},'' 2026. [Online]. Available: \url{https://arxiv.org/abs/2608.20961}
\BIBentrySTDinterwordspacing

\bibitem{ghantasala2026treewycode}
------, ``{TreeWY: Tree Speculative Decoding for Hybrid GDN Models},''
  \url{https://github.com/vllm-project/vllm/issues/54080}, 2026, author
  implementation RFC; accessed September 26, 2026.

\bibitem{replayssm2026}
{Dao AI Lab and NVIDIA}, ``{ReplaySSM: Cache SSM Inputs Instead of State},''
  \url{https://dao-lab.ai/blog/2026/replayssm/}, 2026, author technical report;
  accessed September 26, 2026.

\bibitem{yang2026onela}
\BIBentryALTinterwordspacing
X.~Yang, C.~Peng, Y.~Zhao, L.~Zeng, A.~Hu, J.~Yang, S.~Wang, J.~Lv, X.~Liang,
  C.~Yang, J.~Liu, and Y.~Qiu, ``{OneLA: Scaling Linear-Attention Decoding to
  Large Beams in Generative Recommendation},'' 2026. [Online]. Available:
  \url{https://arxiv.org/abs/2609.12399v1}
\BIBentrySTDinterwordspacing

\bibitem{he2025determinism}
H.~He, ``{Defeating Nondeterminism in LLM Inference},''
  \url{https://thinkingmachines.ai/blog/defeating-nondeterminism-in-llm-inference/},
  2025, {Thinking Machines Lab}, September 10, 2025.

\bibitem{oliaro2025suffix}
\BIBentryALTinterwordspacing
G.~Oliaro, Z.~Jia, D.~Campos, and A.~Qiao, ``{SuffixDecoding: Extreme
  Speculative Decoding for Emerging AI Applications},'' 2025. [Online].
  Available: \url{https://arxiv.org/abs/2411.04975v3}
\BIBentrySTDinterwordspacing

\bibitem{arctic2026suffix}
{Snowflake Arctic Inference}, ``{Suffix Decoding},''
  \url{https://arcticinference.readthedocs.io/en/latest/suffix-decoding.html},
  2026, implementation documentation; accessed September 24, 2026.

\bibitem{hu2024sam}
\BIBentryALTinterwordspacing
Y.~Hu, K.~Wang, X.~Zhang, F.~Zhang, C.~Li, H.~Chen, and J.~Zhang, ``{SAM
  Decoding: Speculative Decoding via Suffix Automaton},'' 2024. [Online].
  Available: \url{https://arxiv.org/abs/2411.10666v3}
\BIBentrySTDinterwordspacing

\bibitem{lumo2026archive}
Z.~Ma, ``{LumoTree: Implementation and Experiment Artifact},''
  \url{https://github.com/MaCoredroid/Lumo_FlyWheel/tree/lumotree-v2-artifact-r2/papers/gdn-tree-scan-mlsys/v2},
  2026, version 2; corrected tree sampler, paper source, and experiment
  records.

\end{thebibliography}
\end{document}